\documentclass[11pt]{article}

\usepackage[margin=1in]{geometry}
\usepackage{times}
\usepackage{natbib}
\setcitestyle{authoryear,round,citesep={;},aysep={,},yysep={;}}

\usepackage{amsmath,amsfonts,bm}

\def\eqref#1{equation~\ref{#1}}

\def\1{\bm{1}}

\DeclareMathAlphabet{\mathsfit}{\encodingdefault}{\sfdefault}{m}{sl}
\SetMathAlphabet{\mathsfit}{bold}{\encodingdefault}{\sfdefault}{bx}{n}

\usepackage{booktabs}
\usepackage{multirow}
\usepackage{enumitem}
\usepackage{amsmath}
\usepackage{amssymb}
\usepackage{marvosym}
\usepackage{graphicx}
\usepackage{tikz}
\usetikzlibrary{arrows.meta}
\usepackage{wrapfig}
\usepackage{float}
\usepackage[table]{xcolor}
\usepackage{subcaption}
\usepackage{array}
\usepackage{tabularx}
\usepackage{url}
\usepackage[hidelinks]{hyperref}
\hypersetup{
  pdftitle={Cartender: Verifiable Language-Conditioned Driving by Learning the Drives Not Taken},
  pdfauthor={Zhiwen Chen, Benjin Zhu, Yan Xie, Jun Ma}
}
\definecolor{tableblue}{RGB}{230,240,250}
\definecolor{tablegray}{HTML}{F0F2F4}
\definecolor{evidencebg}{HTML}{FFFBF0}
\definecolor{evidencegreen}{HTML}{EAF3ED}
\definecolor{branchgreen}{RGB}{238,247,240}

\title{Cartender: Verifiable Language-Conditioned\\Driving by Learning the Drives Not Taken}
\author{
Zhiwen Chen\textsuperscript{1,*} \quad
Benjin Zhu\textsuperscript{2,*,$\bigstar$} \quad
Yan Xie\textsuperscript{2} \quad
Jun Ma\textsuperscript{3,\Letter}\\[0.5em]
{\small $^{1}$The Hong Kong University of Science and Technology (hkust-gz.edu.cn)}\\
{\small $^{2}$Li Auto Inc. (lixiang.com)}\\
{\small $^{3}$The Hong Kong University of Science and Technology (ust.hk)}
}
\date{}

\begin{document}
\maketitle
{\renewcommand{\thefootnote}{}\footnotetext[0]{\textsuperscript{*} Equal contribution.\quad
\textsuperscript{$\bigstar$} Project leader.\quad
\textsuperscript{\Letter} Corresponding author.}}

\begin{abstract}
Every driving scene is logged exactly once, yet driving is inherently one-to-many. 
The scene decides which behaviors are admissible and the driver's intent which one to take: vision alone cannot choose
among them, and language alone cannot tell what the scene permits.
But the log keeps only the drive that was taken, and its instruction merely describes it.
Language-conditioned driving policies are reported to follow instructions, yet on such data this cannot be established:
a policy that ignores the instruction and one that obeys it regardless of the scene fit the logs equally well, and neither is distinguishable from one that weighs both.
Trained only on such logs, driving vision--language--action models (VLAs) and world--action models (WAMs) receive no direct supervision on actions that violate their instructions or instructions the scene rules out.
We present \textbf{Cartender}, which learns from the drives not taken.
To construct matched and mismatched instruction--action pairs from real recordings without a simulator, we propose \emph{Counterfactual Branching},
which treats each logged trajectory as one branch among several candidate futures: a vision--language model proposes the others and the instruction that selects each, and instructions the scene cannot support become negative labels.
Cartender then weighs both explicitly: a learned scene--language compatibility regulates language conditioning during trajectory generation, and a reward-trained language residual re-ranks the candidates over a scene-based score.
Branches serve training; instruction following is evaluated by changing instructions under identical recorded inputs, using behavior criteria rather than generated trajectory targets.
On WOD-E2E, the best of four proposals reaches an RFS of 8.642 across 3,603 feasible instruction conditions.
Across all 8,974 admitted proposals, mean RFS is 8.203 versus 7.906 for their matched logged trajectories, and 67\% score higher than their logged counterpart.
Relative to matched-only imitation with the same backbone,
Cartender more than doubles the rate of fulfilling both instructions of a same-scene pair (57.00\% vs.\ 23.67\%) while RFS rises from 6.32 to 7.10; its gate cuts execution of infeasible instructions from 53.36\% to 14.18\%.
\end{abstract}

\section{Introduction}
\label{sec:intro}

\begin{wrapfigure}[11]{r}{0.47\linewidth}
\vspace{-1.1em}
\centering
\definecolor{bnOK}{HTML}{2E7D32}
\definecolor{bnBad}{HTML}{C62828}
\definecolor{bnReq}{HTML}{1565C0}
\definecolor{bnRoad}{HTML}{DDDDDD}
\begin{tikzpicture}[
    font=\small, line cap=round, >={Stealth[length=4pt]},
    crossing/.pic={
      \fill[bnRoad] (-0.45,-1.2) rectangle (0.45,1.2);
      \fill[bnRoad] (-1.2,-0.45) rectangle (1.2,0.45);
      \draw[white, line width=0.7pt, dash pattern=on 2.5pt off 2.5pt]
        (0,-1.2) -- (0,-0.45) (0,0.45) -- (0,1.2)
        (-1.2,0) -- (-0.45,0) (0.45,0) -- (1.2,0);
      \fill[black!75, rounded corners=1pt] (0.10,-1.12) rectangle (0.34,-0.74);
    }]
  \begin{scope}[yshift=2.07cm]
    \draw[black!30, rounded corners=3pt] (0,0) rectangle (6.3,1.95);
    \node[anchor=west] at (0.1,1.72) {\textbf{Training:} matched pairs only};
    \begin{scope}[shift={(0.95,0.8)}, scale=0.6, transform shape]
      \pic {crossing};
      \draw[->, bnOK, line width=1.2pt] (0.22,-0.72) -- (0.22,1.1);
    \end{scope}
    \node[anchor=west, align=left] at (1.85,1.12)
      {\textcolor{bnOK}{$\checkmark$}~seen~$\langle L, A\rangle$\\
       \emph{``go straight''} $+$ straight};
    \node[anchor=west, align=left, text=black!45] at (1.85,0.42)
      {$\times$~never seen~$\langle L_{\mathrm{neg}}, A\rangle$\\
       \emph{``turn left''} $+$ straight};
  \end{scope}
  \draw[black!30, rounded corners=3pt] (0,0) rectangle (6.3,1.95);
  \node[anchor=west] at (0.1,1.72) {\textbf{Evaluation:} same scene, new instruction};
  \begin{scope}[shift={(0.95,0.8)}, scale=0.6, transform shape]
    \pic {crossing};
    \draw[->, bnReq, line width=1.2pt, dash pattern=on 3pt off 2pt]
      (0.22,-0.72) .. controls (0.22,0.15) and (0,0.22) .. (-1.1,0.22);
    \draw[->, bnBad, line width=1.2pt] (0.22,-0.72) -- (0.22,1.1);
  \end{scope}
  \draw[bnReq, line width=1.2pt, dash pattern=on 3pt off 2pt] (1.95,1.22) -- (2.4,1.22);
  \node[anchor=west] at (2.45,1.22) {$L$: \emph{``turn left''}};
  \draw[bnBad, line width=1.2pt] (1.95,0.78) -- (2.4,0.78);
  \node[anchor=west] at (2.45,0.78) {$A$: drives straight};
  \node[anchor=west, text=bnBad] at (1.9,0.3) {\textbf{MISMATCH}};
\end{tikzpicture}
\vspace{-0.8em}
\end{wrapfigure}
The same driving situation admits many competent behaviors --- overtake the slow truck now or follow it for three seconds first, take the gap at 40\,km/h or at 25, return to the right lane the moment it is clear or stay out until the exit --- and the scene settles only which are admissible, never which one is preferred~\citep{liao2025diffusiondrive,wang2026drivemyway}.
That is settled by an intention the sensors never see.
They record the situation, not the reason one behavior was chosen, and not the behaviors passed over.
Nor can more data recover the rest.
A vehicle occupies one position in space and time and drives one trajectory; the situation never reassembles itself for a second attempt.
Every driving corpus therefore contains single samples from genuinely multi-modal conditional behavior distributions --- one sample each, whatever else is recorded alongside it.
The alternatives are not missing from the datasets we happen to have.
They are not collectible as outcomes of that same event.

\begin{figure}[H]
    \centering
    \includegraphics[
        width=0.85\linewidth,
        pagebox=cropbox
    ]{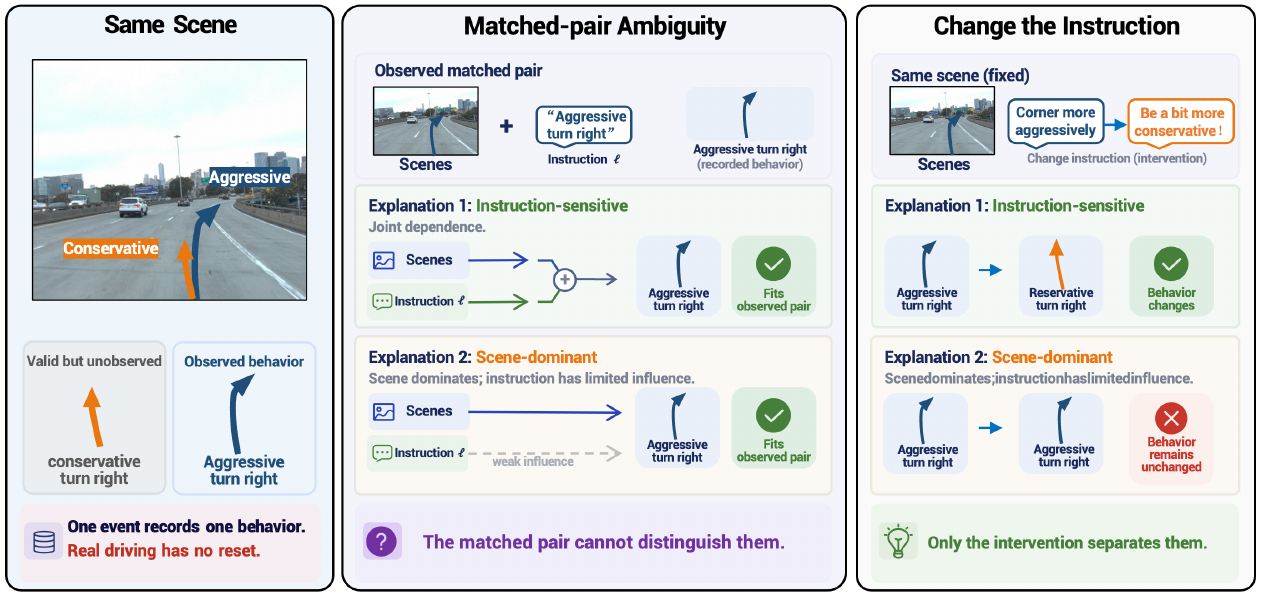}

    \caption{
        \textbf{Matched-pair ambiguity in real driving.}
       A single event reveals one recorded behavior, so the same matched instruction--trajectory pair can be consistent with policies exhibiting different instruction dependence. Holding the scene fixed and changing the instruction separates them.
    }
    \label{fig:matched_pair}
\end{figure}

Their absence is felt twice: once in what can be trained
and, with far less attention, once in what can be checked.
Matched records show the executed behavior, leaving alternative responses unobserved.
Establishing instruction control requires testing what
the policy does under a \emph{different} instruction
in the \emph{same} scene.
Matched instruction--trajectory evaluation leaves this response untested.
The instruction-following scores reported on such data are taken to show that the policy reads the instruction.
They do not.
The visual scene may not determine which admissible behavior is preferred, but it can already narrow the space of plausible actions enough for a largely scene-driven policy to fit the recorded trajectory well; an instruction-sensitive policy can fit the same matched pair just as well, while differing only in what happens when the instruction is changed.
Both score well.
Figure~\ref{fig:matched_pair} makes the ambiguity concrete.
Only an instruction intervention separates them.

\vspace{-0.5cm}

Simulation can construct alternative trajectories from a
recorded simulator state.
SimLingo's Action Dreaming pairs language instructions with alternative trajectories generated from recorded CARLA states through offline, non-reactive simulation~\citep{renz2025simlingo}.
Our construction instead operates on real recordings without a simulator.
Matched-pair evaluation on these recordings retains the ambiguity above.
Whether same-scene supervision helps is not the open question; simulation already indicates that it does.
Two others are.
The first is about data: whether such supervision can be obtained, and trusted, where the situation cannot be replayed.
The second does not wait on the first.
Instruction dependence can already be measured on
released models, and that is where we begin.

Two of the three diagnostics we use to measure instruction dependence need no data beyond what a model was already trained on.
Instructions are first shuffled across scenes. If the action loss does not move, the matched objective is insensitive to instruction identity.
A rise exposes dependence.
At inference, removing the instruction tests whether
its presence changes the predicted trajectory.
Neither diagnostic requires a behavior that was never
driven, so the confound becomes something to measure
rather than something to argue about.
On LMDrive, shuffling increases waypoint L1 loss by 0.011.
Under the evaluated Qwen-Drive interface, instruction
removal shifts predictions by only 0.064\,m.
The third holds the scene fixed and changes the instruction,
probing alternative behaviors that a single recorded
event cannot jointly reveal.

We construct same-scene, multi-intent supervision on real driving data, treating its two halves differently.
Declaring an instruction \emph{inapplicable} to a recorded scene takes a label and no trajectory at all, so scene--instruction applicability is grounded in real observation by construction; supplying the behavior a \emph{feasible} alternative instruction would have produced takes a trajectory that was never driven, and Sec.~\ref{sec:relational_supervision} states how it is branched and on what grounds it is admitted.
We then keep training and measurement apart.
Branched alternatives enter training. Evaluation changes instructions under an identical recorded input, with both requests reviewed for feasibility and predictions assessed by behavior criteria, not by distance to generated teachers.

The feasible half is the hard one: it requires the behavior that a valid alternative instruction would have produced, and we supply it with \emph{Counterfactual Branching}.
At selected keyframes, a vision--language model proposes route-relative parameters and corresponding instructions; deterministic construction and rule-based admission produce trajectory branches in the planner's prediction space.
Branch quality is an empirical question, not a design assumption. We evaluate admitted branches against matched logged trajectories using Rater Feedback Score (RFS).
Across the complete set of 3,603 feasible instruction
conditions, the branched set reaches a best-of-four RFS of 8.642.
Across all 8,974 admitted proposals, mean RFS is 8.203, compared with 7.906 for their matched logged trajectories, and 67\% of branches individually outperform their logged counterparts.

A demonstration identifies a point; a preference among admissible behaviors is an ordering.
That distinction follows from multimodality: a demonstration says ``drive like this,'' but does not rank admissible alternatives under an instruction; a reward expresses that ordering.
The supervision supplies the candidates, not the ordering.
Same-scene alternatives make the comparison expressible;
the ordering itself is defined by the frozen key-event
reward in Sec.~\ref{sec:relation_learning}.
Cartender accordingly generates a candidate set, estimates continuous scene–language compatibility to scale language conditioning during generation and ranking, and ranks the candidates with a demonstration-initialized base score plus a reward-trained language residual.
Our main contributions are:
\begin{enumerate}[leftmargin=1.8em,itemsep=0.2em]

    \item
    \textbf{The matched-pair confound, and how to measure it.}
    An instruction-following score on matched
    instruction--trajectory pairs does not establish
    that the policy's behavior depends on the instruction.
    We expose this ambiguity through instruction removal,
    shuffling, and same-scene swapping.
    Removal and shuffling, where native action loss is
    accessible, need no alternative trajectory targets;
    swapping fixes a recorded input and tests two feasible requests
    without generated trajectory targets.

    \item
    \textbf{Counterfactual Branching.}
    Same-scene, multi-intent supervision built on real
    recordings without a simulator, where the logged
    trajectory anchors route-relative candidate futures,
    each paired with a proposed instruction.
    Branches pass rule-based admission before RFS evaluation:
    admitted branches average 8.203 against 7.906 for
    matched logged trajectories, with 67\% individually
    scoring higher.
    Inapplicable instructions require labels but no
    synthesized trajectories.

    \item
    \textbf{Cartender.}
    A policy that gates language conditioning through
    scene--language compatibility and ranks candidates
    with a reward-trained language residual over a
    demonstration-initialized base score.
    Cartender more than doubles maneuver paired success over matched-only imitation (57.00\% vs.\ 23.67\%),
    with RFS rising from 6.32 to 7.10. Gating reduces unsupported-request following from 53.36\% to 14.18\%.

\end{enumerate}

\section{Related Work}
\label{sec:related_work}

\paragraph{Multi-Proposal Trajectory Planning.}
Multi-proposal planners address behavioral multimodality
by predicting several trajectories per scene.
DiffusionDrive~\citep{liao2025diffusiondrive} and
DiffusionDriveV2~\citep{zou2025diffusiondrivev2} generate candidates
through truncated diffusion and select among them with a learned
scorer.
The candidate set hedges against missing one answer;
it does not establish which answer the instruction prefers.
We supply what matched demonstrations alone do not---a
same-scene criterion for intention-dependent selection.

\paragraph{Driving Vision--Language--Action Models.}
LMDrive~\citep{shao2024lmdrive}, ORION~\citep{fu2025orion},
AutoVLA~\citep{zhou2025autovla}, and
MindVLA-U1~\citep{huang2026mindvla} connect language
with trajectory generation through different architectures.
NudgeVAD~\citep{yang2026nudgevad} adds language-conditioned
residuals to a frozen planner and examines when navigation
commands make language redundant.
The central issue here is not architecture but what
matched-pair training and evaluation establish.
On matched pairs, the instruction and recorded behavior
co-occur; agreement with that behavior alone does not
establish that the prediction changes when the instruction does.
ICR-Drive~\citep{hamid2026icrdrive} tests robustness to
instruction perturbations under matched simulator routes and seeds.
Our question is whether different feasible instructions
select different appropriate behaviors under the same
recorded input.
Our contribution is diagnostic before it is architectural:
we distinguish instruction dependence from agreement with
a matched trajectory, with two tests requiring no new annotation.

\paragraph{Language--Action Alignment and Driving Supervision.}
LinkVLA~\citep{wang2026unifying} strengthens language--action
coupling through shared representations, while Drive My
Way~\citep{wang2026drivemyway} learns driving preferences.
SimLingo's Action Dreaming~\citep{renz2025simlingo} constructs
alternatives through offline, non-reactive simulation;
MALT-Drive~\citep{lu2026maltdrive} uses rollout-based
counterfactual supervision to align control instructions
with planning.
Our setting requires alternative trajectories for real
recordings without simulator rollouts.
OmniDrive~\citep{wang2025omnidrive} supplies counterfactual
descriptions of alternative outcomes; constructing the trajectories
themselves additionally requires checking their validity.
Counterfactual Branching does not sidestep this requirement.
It constructs alternatives from recorded routes, admits
them through rule-based checks, and evaluates their quality
afterward using RFS (Sec.~\ref{sec:relational_supervision});
these checks do not certify full scene feasibility.
Generated teachers serve training, not evaluation targets:
instruction control is assessed by behavior criteria under
fixed recorded inputs.

\section{Method}
\label{sec:method}

\begin{figure}[t]
\centering
\includegraphics[width=\linewidth]{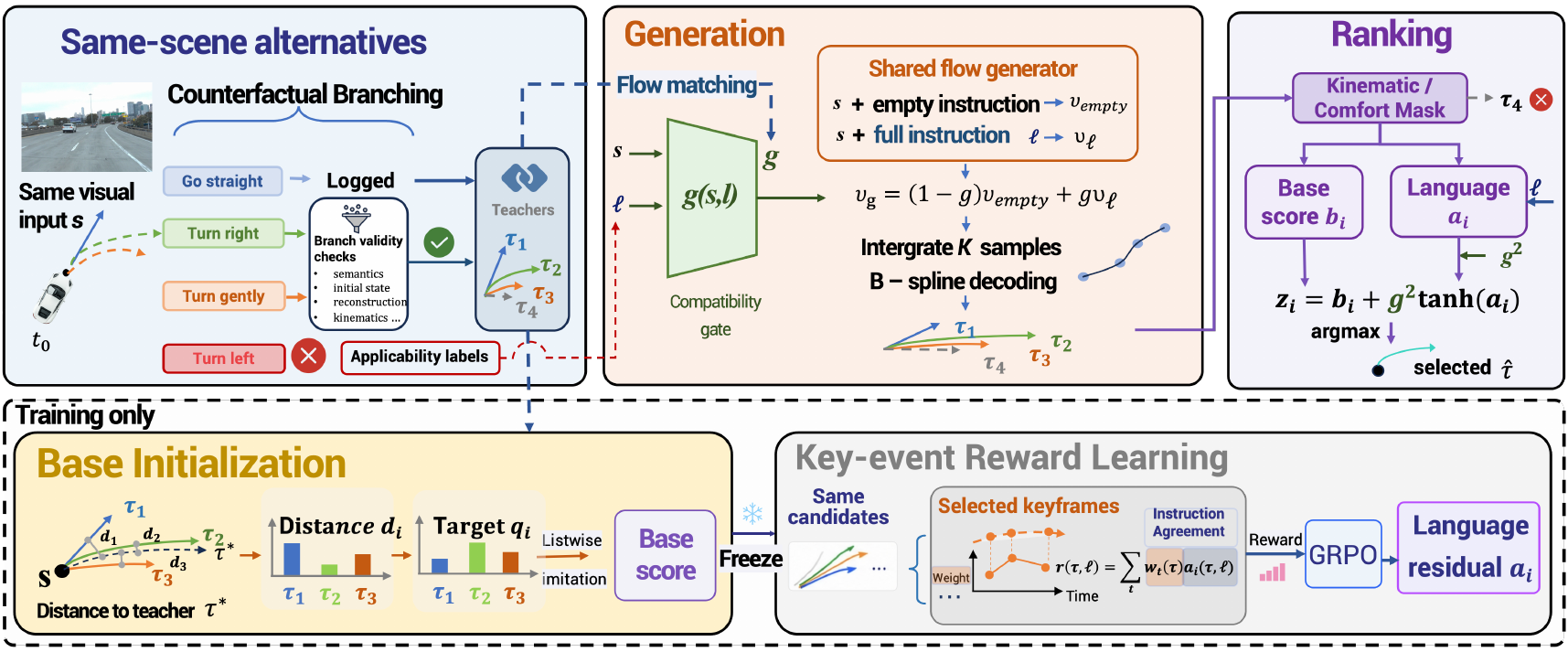}
\caption{
Overview of Cartender.
Offline same-scene relations provide applicability labels for
compatibility learning and teacher trajectories for flow matching
and base-score initialization.
Compatibility modulates generation and a language-dependent
ranking residual.
After base initialization, reward learning updates only the residual.
At inference, the policy ranks candidates passing kinematic and
comfort checks without teachers or reward evaluation.
}
\label{fig:overview}
\end{figure}

Given a visual scene $s$ and an instruction $\ell$, Cartender
predicts an ego trajectory $\tau\in\mathbb{R}^{T\times2}$.
Same-scene relations supply two signals: which requests the scene
supports, and which trajectories realize them.
Figure~\ref{fig:overview} shows how these signals govern generation
and candidate ranking.

\subsection{Same-Scene Relational Supervision}
\label{sec:relational_supervision}

A single visual scene can support multiple distinct driving
behaviors while leaving other requested behaviors unsupported.
We organize supervision around this shared context: feasible
instructions receive teacher trajectories, and infeasible
instructions receive negative applicability labels.
The two halves are not equally hard to obtain.
An infeasible relation needs no counterfactual trajectory;
a feasible alternative requires behavior absent from the recording.
Counterfactual Branching supplies it.
At selected keyframes, a vision--language model proposes alternative
futures and their instructions alongside the logged branch.
We reconstruct proposals in the planner's continuous trajectory
space, enforcing observed initial position and velocity before
boundary-preserving smoothing. We check source validity, semantic conflicts, initial-state
consistency, reconstruction error, and kinematic and comfort
violations, including those after reconstruction.
Passing these checks does not establish driving quality.
Section~\ref{sec:exp_branching} evaluates the best branches
and admitted set using automated Rater Feedback Score (RFS);
RFS does not determine admission.

Branched trajectories supervise training. The diagnostic in
Sec.~\ref{sec:exp_swap} reuses one recorded input snapshot for two
instructions independently reviewed as feasible. It does not pair
approximately similar scenes or assume both responses were recorded.
Behavior criteria assess predictions without generated trajectory targets.

Binary applicability labels train a continuous compatibility score.
Let $q_{\ell}(k\mid\ell)$ be the frozen language-intent posterior
and $p_s(k\mid s)$ the scene-support probability for intent $k$.
The requested intent $\hat{k}$ and confidence $c_{\ell}$ define
\begin{equation}
\hat{k}=\arg\max_k q_{\ell}(k\mid\ell),
\qquad
c_{\ell}=\max_k q_{\ell}(k\mid\ell),
\qquad
g(s,\ell)=c_{\ell}\,p_s(\hat{k}\mid s).
\label{eq:compatibility}
\end{equation}
We train support with intent-balanced binary cross-entropy
on feasible and infeasible relations, then freeze the compatibility
model. The language-intent branch remains fixed.
The gate estimates maneuver-level support; the generator and ranking residual condition on the full instruction, including timing, speed, and manner.
Section~\ref{sec:exp_swap} evaluates maneuver and qualifier control separately.
\subsection{Learning from Structured Language--Behavior Relations}
\label{sec:relation_learning}

Teacher trajectories supervise conditional flow matching in
B-spline coefficient space; a fixed decoder $D_{\mathrm{BS}}$
maps coefficients $c$ to $T$-step trajectories~\citep{han2026bspline}.
At flow time $t$ and state $c_t$, the same generator produces
empty-instruction and full-instruction fields.
Compatibility sets their classifier-free guidance weight:
\begin{equation}
v_g(c_t,t)
=
v_{\emptyset}(c_t,t;s)
+
g(s,\ell)
\left[
v_{\ell}(c_t,t;s,\ell)
-
v_{\emptyset}(c_t,t;s)
\right].
\label{eq:gate_flow}
\end{equation}
With $g\in[0,1]$, scene–language compatibility interpolates between the scene-conditioned empty-instruction field and the full-instruction field, controlling the strength of language conditioning.
Integrating from $K$ noise samples and decoding yields
$\mathcal{T}(s,\ell)=\{\tau_i\}_{i=1}^{K}$.

We mask candidates violating kinematic or comfort constraints
and denote the surviving indices by $\mathcal{A}$.
Their logits combine an imitation-initialized base with a
bounded language residual:
\begin{equation}
z_i
=
b_{\psi}(s,\tau_i)
+
g(s,\ell)^2
\tanh a_{\theta}(s,\ell,\tau_i).
\label{eq:factorized_score}
\end{equation}
The base reads the scene and trajectory; the residual also reads
the instruction. Scaling by $g^2$ limits language-driven ranking
changes at low compatibility.
Inference selects the highest-scoring candidate under a masked softmax;
if none survives, it returns no trajectory.

Geometric proximity to a teacher need not imply instruction fulfillment.
We therefore initialize ranking by teacher proximity and refine it
with the key-event reward.
For each feasible relation $(s,\ell)$, proximity to its admitted
A and B teachers $\mathcal{H}(s,\ell)$ defines a target distribution
$q_i$ over $i\in\mathcal{A}$.
C and D are not teacher targets.
We train the base scorer with
\begin{equation}
\mathcal{L}_{\mathrm{IL}}
= -\sum_{i\in\mathcal{A}} q_i\log\pi_{\psi}(i\mid s,\mathcal{T}),
\label{eq:listwise_il}
\end{equation}
averaged over feasible relations, where $\pi_{\psi}$ is the
masked softmax of base scores.
During base initialization, the language residual stays fixed at zero.

We then freeze the generator, compatibility model, and base scorer.
Only the language residual learns, through Group Relative Policy
Optimization (GRPO)~\citep{shao2024deepseek}.
A frozen key-event evaluator supplies the reward
\begin{equation}
r(\tau,\ell)
=
\sum_{t=1}^{T}w_t(\tau)\,a_t(\tau,\ell),
\qquad
\sum_{t=1}^{T}w_t(\tau)=1,
\label{eq:event_reward}
\end{equation}
where $t$ indexes trajectory steps, $w_t$ emphasizes salient motion
changes, and $a_t$ measures trajectory-observable instruction
agreement. GRPO normalizes relative advantages within each
scene--instruction group before weighting reward-driven updates
by $g^2$. Low-compatibility groups receive a stronger anchor to
the frozen imitation policy.

Finite candidates and frozen rewards also permit exact summation
of the selector's expected reward over unmasked candidates.
GRPO is therefore an optimization choice rather than a requirement
of the formulation.
Section~\ref{sec:exp_ablation} compares both optimizers and tests
gating, preference rewards, and staged optimization.
Teachers and reward evaluation are training-only.
Low compatibility attenuates language influence in generation
and ranking; it does not produce an explicit refusal.

\section{Experiments}
\label{sec:experiments}

\subsection{Experimental Setup}
\label{sec:exp_setup}

\paragraph{Data and comparisons.}
Our data pool comprises 10,336 training and 4,580 validation
scenes from WOD-E2E~\citep{xu2025wode2ewaymo}; the quality
evaluation uses 1,200 validation scenes.
Unless otherwise stated, Cartender variants and the
language-free control share frozen U1 scene features,
a 5\,s horizon, and $K=24$ candidates.
Released-policy diagnostics use fixed checkpoints of
Qwen-Drive~\citep{zhou2026qwendrive},
SimLingo~\citep{renz2025simlingo}, and
LMDrive~\citep{shao2024lmdrive}:
Qwen-Drive uses our added instruction interface on WOD-E2E;
SimLingo and LMDrive use their respective CARLA protocols.

\paragraph{Supervision and training.}
Counterfactual Branching constructs 3,603 feasible instruction
conditions across 2,487 source scenes.  Four proposals per
condition yield 14,412 trajectories, of which 8,974 pass
rule-based admission.  The admitted A and B proposals form the
teacher sets; all 3,603 A proposals pass admission, while C and D
are excluded from teacher supervision.  Infeasible instructions
supply applicability labels only.
Selector comparisons use three seeds with a shared
generator checkpoint, candidate sets, and fixed reward
(Sec.~\ref{sec:relation_learning});
Table~\ref{tab:main_ablation} reports mean paired success.

\paragraph{Evaluation.}
We test instruction dependence through removal,
shuffling where native action loss is accessible,
and same-scene swapping between feasible instructions.
Maneuver-level and qualifier-level results are reported separately.
The maneuver diagnostic uses 200 pairs; the qualifier diagnostic
uses 64 speed pairs and 32 timing pairs.
Paired success requires both requests to satisfy action and hard-constraint checks.
Behavioral criteria are independent of Cartender's
relational training annotations.
We also report RFS, ADE, and output coverage.
Missing predictions count as failures in success metrics;
quality statistics state their coverage.

\subsection{Is the Instruction Being Read at All?}
\label{sec:exp_diagnosis}

Table~\ref{tab:instruction_dependence} tests frozen released models
under their respective input interfaces and datasets, without our
constructed supervision. These are within-model interventions,
not a common cross-model benchmark; Qwen-Drive results are specific
to the added instruction interface.
We evaluate instruction dependence using three controlled interventions:
instruction removal, instruction shuffling, and same-scene instruction swapping.
\begin{table*}[t]
\centering
\caption{Instruction dependence and same-scene control.}
\label{tab:instruction_diagnostics}

\begingroup
\small
\setlength{\tabcolsep}{0pt}
\setlength{\extrarowheight}{0pt}
\renewcommand{\arraystretch}{1}
\captionsetup[subtable]{font=small,justification=centering,singlelinecheck=false,skip=2pt}

\newcommand{\ABrowheight}{26.66667pt}
\newcommand{\ABcell}[2]{\parbox[c][\ABrowheight][c]{\linewidth}{\centering #2}}
\newcommand{\ABname}[2]{\parbox[c][\ABrowheight][c]{\linewidth}{\raggedright #2}}
\newcommand{\ABhead}[1]{\parbox[c][44pt][c]{\linewidth}{\centering\bfseries #1}}

\begin{subtable}[t]{0.485\linewidth}
\centering
\caption{Released-policy diagnostics.}
\label{tab:instruction_dependence}
\small
\begin{tabular}{@{}p{0.22\linewidth}p{0.14\linewidth}p{0.20\linewidth}p{0.20\linewidth}p{0.24\linewidth}@{}}
\toprule
\rowcolor{tablegray}
\ABhead{Model} &
\ABhead{Ref.\\ADE\\(m)} &
\ABhead{Removal\\shift\\(m)} &
\ABhead{Shuffle\\$\Delta\mathcal{L}_{\mathrm{action}}$} &
\ABhead{Swap\\paired\\success (\%)} \\
\midrule
\ABname{20pt}{Qwen-\\Drive} &
\ABcell{20pt}{1.413} &
\ABcell{20pt}{0.064} &
\ABcell{20pt}{\textemdash} &
\ABcell{20pt}{21.13} \\
\ABname{20pt}{SimLingo} &
\ABcell{20pt}{0.842} &
\cellcolor{evidencegreen}\ABcell{20pt}{2.511} &
\ABcell{20pt}{$+0.232$\\$+1.289$} &
\cellcolor{evidencegreen}\ABcell{20pt}{50.00} \\
\ABname{20pt}{LMDrive} &
\cellcolor{evidencegreen}\ABcell{20pt}{0.203} &
\ABcell{20pt}{0.178} &
\ABcell{20pt}{$+0.011$} &
\cellcolor{evidencegreen}\ABcell{20pt}{3.20} \\
\bottomrule
\end{tabular}
\end{subtable}\hfill
\begin{subtable}[t]{0.485\linewidth}
\centering
\caption{Same-scene instruction control.}
\label{tab:same_scene_swap}
\small
\renewcommand{\ABrowheight}{22pt}
\begin{tabular}{@{}p{0.24\linewidth}p{0.25\linewidth}p{0.245\linewidth}p{0.265\linewidth}@{}}
\toprule
\rowcolor{tablegray}
\ABhead{Model} &
\ABhead{Maneuver\\success\\(\%)} &
\ABhead{Trajectory\\shift (m)} &
\ABhead{Direction\\consistency\\(\%)} \\
\midrule
\ABname{15pt}{Vision-only} &
\ABcell{15pt}{20.37} &
\ABcell{15pt}{0.00} &
\ABcell{15pt}{0.00} \\
\ABname{15pt}{Qwen-\\Drive} &
\ABcell{15pt}{27.78} &
\ABcell{15pt}{0.60} &
\ABcell{15pt}{22.22} \\
\parbox[c][14pt][c]{\linewidth}{\raggedright SimLingo} &
\parbox[c][14pt][c]{\linewidth}{\centering 72.80} &
\parbox[c][14pt][c]{\linewidth}{\centering 0.89} &
\parbox[c][14pt][c]{\linewidth}{\centering 62.50} \\
\ABname{15pt}{\textbf{Cartender}} &
\cellcolor{evidencegreen}\ABcell{15pt}{\textbf{73.83}\\\textbf{$\pm$\,2.60}} &
\ABcell{15pt}{$1.666$\\$\pm0.09$} &
\cellcolor{evidencegreen}\ABcell{15pt}{\textbf{83.09}\\\textbf{$\pm$\,7.40}} \\
\bottomrule
\end{tabular}
\end{subtable}

\par\vspace{2pt}
\begin{minipage}{\linewidth}
\small\raggedright
\textbf{(a)} Frozen checkpoints follow their protocols.
Ref. ADE uses original instructions; Removal uses empty inputs.
Shuffle reports shuffled-minus-original native action loss,
with scenes and targets fixed.
SimLingo: route loss above speed-waypoint loss; LMDrive: waypoint L1.
Dash: unavailable; shading: within-model contrasts.
Paired success requires both requests fulfilled.
\par
\textbf{(b)} Scenes and random draws are fixed.
Maneuver success measures individual request fulfillment;
trajectory shift measures prediction displacement;
direction consistency measures alignment with the requested direction.
\end{minipage}
\endgroup
\end{table*}

Table~\ref{tab:instruction_dependence} exposes the gap
between predicting the recorded trajectory and following
the requested behavior.
Under the evaluated Qwen-Drive interface, instruction
removal changes predictions by only 0.064\,m, while
same-scene paired success is 21.13\%.
LMDrive attains a reference ADE of 0.203\,m but only
3.20\% paired success under its evaluated protocol.
These results show that low matched trajectory error
does not establish reliable instruction control.
SimLingo clearly uses the instruction: removal shifts
predictions by 2.511\,m, and shuffling increases both
native route and speed-waypoint losses.
Nevertheless, paired success is 50.00\%, leaving at
least one request unfulfilled in half of the pairs.
Using language does not ensure that the requested
behavior is realized; instruction dependence and
behavioral fulfillment must be evaluated separately.

\subsection{Instruction Dependence on Real Same-Scene Pairs}
\label{sec:exp_swap}
The diagnostic reuses recorded input snapshots;
generated teachers are not evaluation targets.
We fix observations and sampling noise while comparing
two instructions feasible under the same observation.
Table~\ref{tab:same_scene_swap} shows that the vision-only control fulfills 20.37\% of requests without changing its output;
fulfillment alone does not establish instruction control. Cartender achieves 83.09\% directional consistency.
The response extends to execution intensity within
the same maneuver.
For gradual versus strong acceleration or deceleration,
speed-direction accuracy is 64.58\% versus 43.75\%;
requiring the correctly directed speed difference
to exceed $0.2\,\mathrm{m/s}$ still yields 53.13\%
success versus 12.50\%.
Timing control remains limited, with 35.42\% correct
early--late ordering across 32 deceleration pairs.

\subsection{Counterfactual Branching: Branch Quality}
\label{sec:exp_branching}

We evaluate constructed branches using RFS, which does not
determine admission.
Best-of-four RFS measures the best available candidate among
A--D for each feasible instruction condition.
Mean admitted RFS and the fraction exceeding matched logged
RFS summarize all admitted proposals, not only the A/B
teachers used for training.
\definecolor{branchgreen}{RGB}{238,247,240}

\begin{table*}[t]
\centering
\normalsize

\caption{
Counterfactual branch quality and validation.
RFS is evaluated after rule-based admission.
Best-of-four averages conditions; mean admitted RFS and
``Above logged'' pool admitted proposals.
``Above logged'' is the fraction exceeding matched logged RFS.
Two reviewers audit constructed infeasible relations.
}
\label{tab:branch_validation}

\vspace{1mm}

\begingroup
\normalsize
\renewcommand{\arraystretch}{1}
\setlength{\extrarowheight}{0pt}
\setlength{\tabcolsep}{1pt}

\begin{minipage}[t]{0.33\linewidth}
\vspace{0pt}
\centering
\begin{tabular*}{\linewidth}{@{\extracolsep{\fill}}lr@{}}
\toprule
\rowcolor{tablegray}
\multicolumn{2}{c}{\textbf{Branch quality}} \\
\midrule
Matched logged RFS       & 7.906 \\
MindVLA-U1 ref. RFS & 7.92  \\
Best-of-4 RFS            & 8.642 \\
\rowcolor{branchgreen}
Mean admitted RFS        & 8.203 \\
\rowcolor{branchgreen}
Above logged (\%)        & 67.0  \\
\bottomrule
\end{tabular*}
\end{minipage}%
\hfill
\begin{minipage}[t]{0.32\linewidth}
\vspace{0pt}
\centering
\begin{tabular*}{\linewidth}{@{\extracolsep{\fill}}lr@{}}
\toprule
\rowcolor{tablegray}
\multicolumn{2}{c}{\textbf{Construction set}} \\
\midrule
Source scenes       & 2,487  \\
Feasible conditions & 3,603  \\
Generated proposals & 14,412 \\
\rowcolor{branchgreen}
Admitted proposals  & 8,974  \\
Admission rate (\%) & 62.27  \\
\bottomrule
\end{tabular*}
\end{minipage}%
\hfill
\begin{minipage}[t]{0.32\linewidth}
\vspace{0pt}
\centering

\newcommand{\auditrowstrut}{%
    \rule[-5.5pt]{0pt}{18.33333pt}%
}

\begin{tabular*}{\linewidth}{@{\extracolsep{\fill}}lr@{}}
\toprule
\rowcolor{tablegray}
\multicolumn{2}{c}{\textbf{Negative-relation audit}} \\
\midrule

\auditrowstrut Audited relations &
228 \\

\auditrowstrut Confirmed infeasible &
202 \\

\rowcolor{branchgreen}
\auditrowstrut Confirmation rate (\%) &
88.6 \\

\bottomrule
\end{tabular*}
\end{minipage}

\endgroup
\end{table*}

Table~\ref{tab:branch_validation} reports two aggregation units.
Best-of-four takes the maximum among A--D for each of 3,603
feasible conditions before averaging.  Mean admitted RFS and
``Above logged'' pool all 8,974 rule-admitted proposals, with
each proposal compared against its condition's logged trajectory.
Under this proposal-level aggregation, admitted branches average
8.203 RFS, exceeding the matched logged score of 7.906 by 0.297.
Moreover, 67\% score above their corresponding logged
trajectories.
The RFS advantage over matched logged trajectories
therefore extends to a majority of admitted branches.
These statistics use the complete construction across 2,487
source scenes rather than a sampled subset.
A separate audit has two reviewers assess each of 228
constructed infeasible relations independently.  Cases with
different judgments receive an additional review.  The audit
confirms 202 relations (88.6\%); 26 remain unconfirmed.

\subsection{Main Results and Elimination of Alternative Explanations}
\label{sec:exp_ablation}

We first evaluate whether instruction control can be improved while preserving trajectory quality.
Beyond this evaluation, we analyze whether the gains can be attributed to additional same-scene examples, training schedules, reward optimization choices, or specific supervision designs.

\textbf{Instruction control and trajectory quality.}
Table~\ref{tab:main_ablation} shows that Cartender more than
doubles maneuver paired success over matched-only imitation
with the same backbone, from 23.67\% to 57.00\%, while RFS
rises from 6.32 to 7.10.
Against the language-free control, maneuver and qualifier
paired success rise from zero to 57.00\% and 35.07\%,
respectively.
RFS rises from 6.24 to 7.10, while output coverage falls
from 98.98\% to 97.70\%.
Paired success requires both requests to be fulfilled;
missing outputs count as failures.

\textbf{Contribution beyond additional trajectory examples.}
Table~\ref{tab:main_ablation} compares matched-only and
same-scene imitation under the same imitation procedure
and optimization budget.
Adding admitted A/B teachers raises maneuver paired success
from 23.67\% to 39.50\%, a gain of 15.83 percentage points.
On the same expanded training set, Cartender further
improves maneuver and qualifier paired success by
17.50 and 26.04 points, reaching 57.00\% and 35.07\%.
Its RFS is 7.10 compared with 7.04 for same-scene
imitation, while coverage increases from 94.78\%
to 97.70\%.
The full training procedure therefore improves
instruction fulfillment beyond imitation on the
additional examples.

\textbf{Compatibility and preference reward.}
Table~\ref{tab:main_ablation} shows that adding
compatibility to same-scene imitation raises maneuver
and qualifier paired success from 39.50\% and 9.03\%
to 45.83\% and 12.85\%, respectively, while RFS decreases
from 7.04 to 6.95.
The full method reaches 57.00\% and 35.07\%, with RFS
increasing to 7.10.
Removing compatibility from the full method reduces
maneuver paired success by 15.33 percentage points
to 41.67\%, eliminates qualifier paired success,
and lowers RFS from 7.10 to 7.00 and coverage from
97.70\% to 94.40\%.
Replacing reward refinement with language-conditioned
residual imitation, using the same initialization and update
count, reduces maneuver and
qualifier paired success by 7.00 and 30.56 points,
respectively, and lowers RFS from 7.10 to 6.37.
Both components therefore contribute to instruction
fulfillment, with compatibility also supporting
output availability.

\textbf{Staged versus joint optimization.}
Both variants start from the same IL endpoint;
the joint variant additionally retains the imitation
loss during preference refinement.
Table~\ref{tab:main_ablation} shows that joint optimization
and staged optimization both achieve 57.00\% maneuver paired
success.  Staged optimization achieves 35.07\% qualifier paired
success versus 5.21\% for joint optimization, and RFS
of 7.10 versus 6.94.
Staged optimization therefore favors qualifier
fulfillment and trajectory quality without changing
maneuver paired success in the reported runs.

\textbf{GRPO versus exact expected-reward optimization.}
We compare GRPO with exact expected-reward optimization
over the same finite candidate set, using the same
reward and imitation anchor.
Table~\ref{tab:main_ablation} shows similar maneuver paired
success: 57.00\% for GRPO and 58.17\% for exact expectation.
GRPO achieves 35.07\% qualifier paired success, compared
with 4.86\% for exact expectation, and RFS of 7.10
versus 7.00.
This fixed-budget result favors GRPO on qualifier fulfillment,
but does not isolate a general optimizer advantage.

\begin{table*}[t]
\centering
\begingroup
\captionsetup{font=normalsize,skip=4pt}
\caption{
Instruction control, trajectory quality, and output coverage.
All methods use logged data and admitted A/B teachers except matched-only imitation.
Imitation baselines share procedure and budget; exact expectation shares
GRPO's candidates, reward, and imitation anchor.
Paired success averages three seeds and requires both requests to succeed;
missing outputs fail. Coverage measures output availability.
}
\label{tab:main_ablation}

\normalsize
\setlength{\tabcolsep}{0pt}
\renewcommand{\arraystretch}{1}
\begin{tabular}{@{}p{0.46\linewidth}*{2}{>{\centering\arraybackslash}p{0.15\linewidth}}>{\centering\arraybackslash}p{0.10\linewidth}>{\centering\arraybackslash}p{0.14\linewidth}@{}}
\toprule
\rowcolor{tablegray}
\textbf{Method} &
\shortstack{\textbf{Maneuver}\\\textbf{paired (\%)} $\uparrow$} &
\shortstack{\textbf{Qualifier}\\\textbf{paired (\%)} $\uparrow$} &
\textbf{RFS} $\uparrow$ &
\shortstack{\textbf{Coverage}\\\textbf{(\%)} $\uparrow$} \\
\midrule
Language-free control           & 0.00  & 0.00 & 6.24 & 98.98 \\
Matched-only imitation          & 23.67 & 3.13 & 6.32 & 96.40 \\
Same-scene imitation            & 39.50 & 9.03 & 7.04 & 94.78 \\
\midrule
Same-scene imitation + compatibility & 45.83 & 12.85 & 6.95 & 97.70 \\
\rowcolor{evidencegreen}
\textbf{Cartender}              & 57.00 & 35.07 & 7.10 & 97.70 \\
\midrule
Cartender, no compatibility     & 41.67 & 0.00 & 7.00 & 94.40 \\
Cartender, no preference reward & 50.00 & 4.51 & 6.37 & 97.70 \\
Joint IL + preference           & 57.00 & 5.21 & 6.94 & 97.70 \\
Exact expected reward           & 58.17 & 4.86 & 7.00 & 97.70 \\
\bottomrule
\end{tabular}
\endgroup
\end{table*}

\begin{figure}[H]
\centering

\begin{subfigure}[t]{0.48\linewidth}
    \centering
    \includegraphics[height=3.4cm,keepaspectratio]{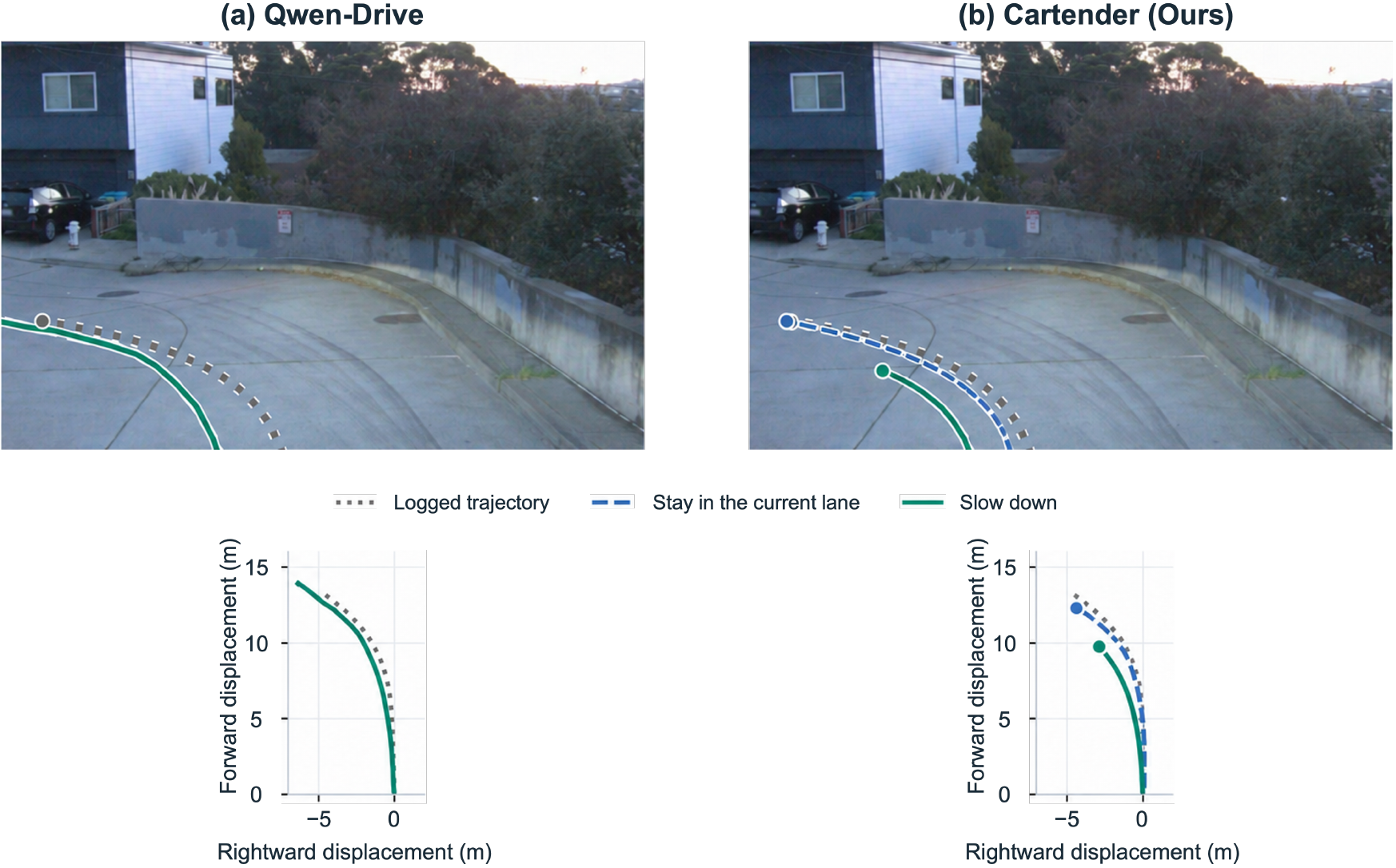}
    \caption{Same-scene instruction intervention.}
    \label{fig:case_same_scene}
\end{subfigure}
\hfill
\begin{subfigure}[t]{0.48\linewidth}
    \centering
    \includegraphics[height=3.4cm,keepaspectratio]{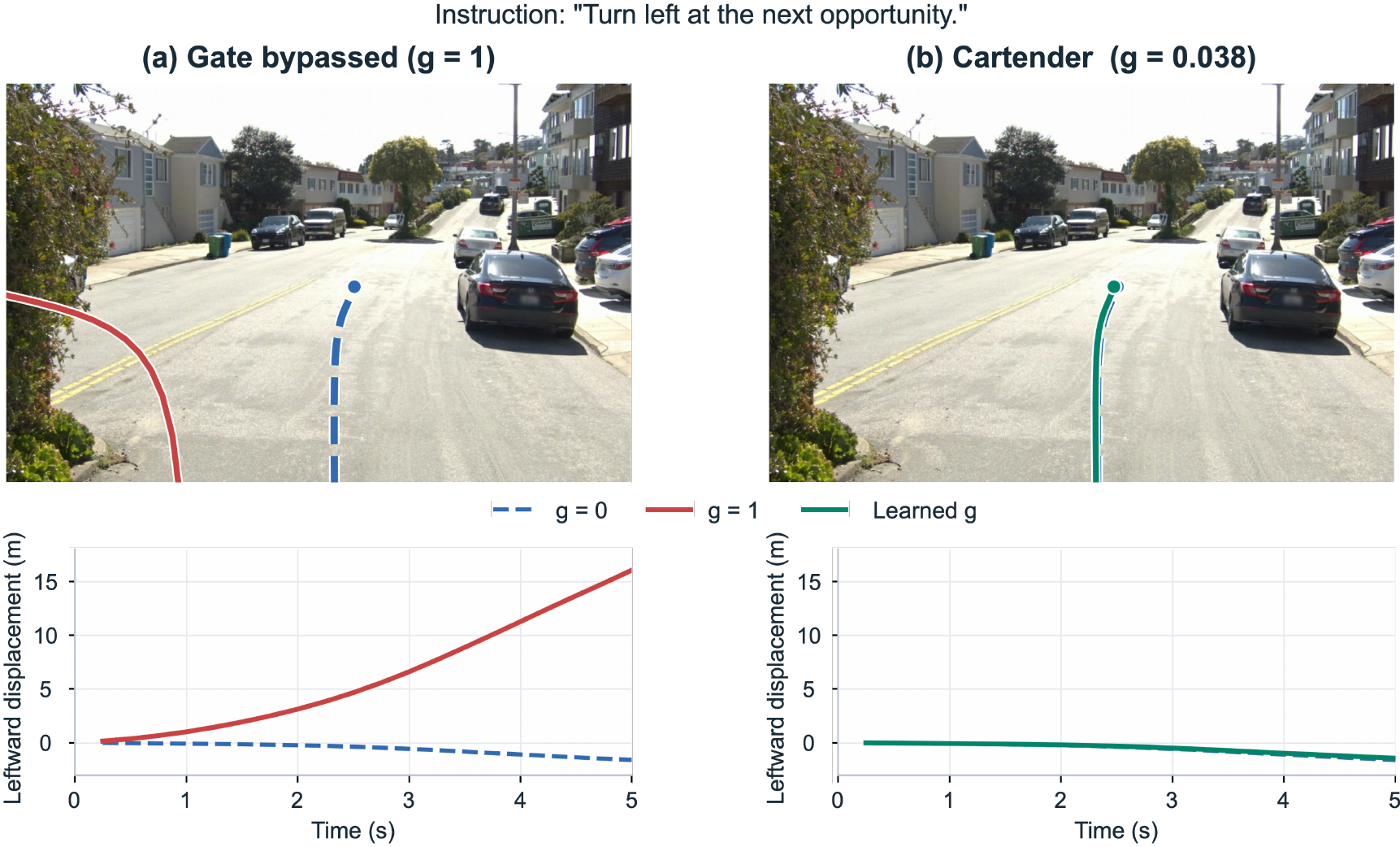}
    \caption{Compatibility gating under an unsupported instruction.}
    \label{fig:case_infeasible}
\end{subfigure}

\caption{
Cartender responds to a supported slow-down request
under fixed scene inputs (left) and suppresses an
unsupported turn through compatibility gating (right).
}
\label{fig:altdrive_cases}
\end{figure}
\paragraph{Qualitative instruction interventions.}

In Fig.~\ref{fig:case_same_scene}, Qwen-Drive predictions
nearly overlap for lane keeping and slowing down,
whereas Cartender predicts a shorter trajectory
for slowing down.
In Fig.~\ref{fig:case_infeasible}, bypassing the gate
produces pronounced leftward motion, while the learned
gate keeps the prediction close to the $g=0$ reference
when the turn is unsupported within the evaluated horizon.

\subsection{Behavior Under Infeasible Instructions}
\label{sec:exp_infeasible}

We test whether scene--language compatibility
distinguishes unsupported instructions and reduces
their influence on predicted behavior.

\paragraph{Compatibility separation.}
Mean compatibility scores are 0.771 for feasible
instructions and 0.310 for infeasible instructions,
with overlapping distributions.
The score achieves an AUROC of 0.863 for identifying
infeasible instructions.
At $\theta=0.5$, classifying low-compatibility instructions
as infeasible yields 76.74\% precision, 74.16\% recall,
and 75.43\% F1.

\paragraph{Behavioral suppression of unsupported instructions.}

With checkpoint, inputs, and random draws fixed, learned
compatibility reduces infeasible execution from 53.36\% with
$g=1$ to 14.18\%, and drift from 3.612 to 0.337\,m.
A separate request-level diagnostic defines Feasible Follow
as valid, hard-constraint-passing fulfillment of feasible requests,
and Relation Handling as its equal-weight average with valid,
hard-constraint-passing nonexecution on infeasible requests.
From no gate to learned compatibility, Feasible Follow falls
from 61.6\% to 43.8\%, while Relation Handling rises from
54.1\% to 64.8\% (61.9\% with a constant gate).
These request-level metrics differ from paired success.
Gating therefore trades feasible-request fulfillment for reduced
unsupported execution in that diagnostic; it provides no explicit refusal.

\section{Conclusion}
\label{sec:conclusion}

Logged driving records one behavior even when several are
admissible, so matched pairs provide neither a direct comparison
among feasible behaviors nor evidence that language caused the
selected one.
Counterfactual Branching constructs same-scene alternatives
with rule-based admission and post-hoc RFS evaluation.
These relations supervise training, while fixed-scene
interventions on recorded inputs use behavior criteria to test
whether predictions follow the changed instruction.

Cartender generates multiple candidates, gates language
conditioning by continuous scene--language compatibility,
and ranks them with a GRPO-trained language residual.
It achieves 57.00\% maneuver paired success versus 23.67\%
for matched-only imitation; gating reduces unsupported-request
following from 53.36\% to 14.18\%.
These results use fixed recorded inputs: rule-based admission and
RFS do not certify collision freedom or instruction fulfillment,
and the gate does not eliminate unsupported execution.
Future work will test continuous instruction updates in
closed-loop, interactive traffic.

\bibliographystyle{plainnat}
\bibliography{references}

\end{document}